\documentclass[11pt,a4paper]{article}
\usepackage[hyperref]{acl2021}
\usepackage{times}
\usepackage{latexsym}
\usepackage{graphicx}
\usepackage{booktabs}
\usepackage{multirow}
\usepackage{xcolor}
\usepackage{float}

\usepackage{microtype}

\aclfinalcopy 

\title{Can I Be of Further Assistance? Using Unstructured Knowledge Access to Improve Task-oriented Conversational Modeling}

\author{Di Jin \\
  Amazon Alexa AI \\
  \texttt{djinamzn@amazon.com} \\\And
  Seokhwan Kim \\
  Amazon Alexa AI \\
  \texttt{seokhwk@amazon.com} \\\And
  Dilek Hakkani-Tur \\
  Amazon Alexa AI \\
  \texttt{hakkanit@amazon.com}\\}

\date{}

\begin{document}
\maketitle
\begin{abstract}
 Most prior work on task-oriented dialogue systems are restricted to limited coverage of domain APIs. However, users oftentimes have requests that are out of the scope of these APIs. This work focuses on responding to these beyond-API-coverage user turns by incorporating external, unstructured knowledge sources. Our approach works in a pipelined manner with knowledge-seeking turn detection, knowledge selection, and response generation in sequence. We introduce novel data augmentation methods for the first two steps and demonstrate that the use of information extracted from dialogue context improves the knowledge selection and end-to-end performances. Through experiments, we achieve state-of-the-art performance for both automatic and human evaluation metrics on the DSTC9 Track 1 benchmark dataset, validating the effectiveness of our contributions. 
\end{abstract}

\section{Introduction}

Driven by the fast progress of natural language processing techniques, we are now witnessing a variety of task-orientated dialogue systems being used in daily life. 
These agents traditionally rely on pre-defined APIs to complete the tasks that users request~\cite{williams:2017,eric-manning:2017}; however, some user requests are related to the task domain but  beyond these APIs' coverage~\cite{kim-etal-2020-beyond}. For example, while task-oriented agents can help users book a hotel,
they fall short of answering potential follow-up questions users may have, such as ``whether they can bring their pets to the hotel''.
These beyond-API-coverage user requests frequently refer to the task or entities that were discussed in the prior conversation and can be addressed by interpreting them in context and retrieving relevant domain knowledge from web pages, for example, from textual descriptions and frequently asked questions (FAQs).
Most task-oriented dialogue systems do not incorporate these external knowledge sources into dialogue modeling, 
making conversational interactions inefficient.  

To address this problem, \citet{kim-etal-2020-beyond} recently introduced a new challenge on task-oriented conversational modeling with unstructured knowledge access, and provided datasets that are annotated for three related sub-tasks: (1) knowledge-seeking turn detection, (2) knowledge selection, and (3) knowledge-grounded response generation (one data sample is in Section B.1 of Supplementary Material).
This problem was intensively studied as the main focus of the DSTC9 Track 1~\cite{kim2020beyond}, where a total of 105 systems developed by 24 participating teams were benchmarked.


In this work, we also follow a pipelined approach and present novel contributions for the three sub-tasks: (1) For knowledge related turn detection, we propose a data augmentation strategy that makes use of available knowledge snippets. (2) For knowledge selection, we propose an approach that makes use of information extracted from the dialogue context via domain classification and entity tracking before knowledge ranking. (3) For the final response generation, we leverage powerful pre-trained models for knowledge grounded response generation in order to obtain coherent and accurate responses. Using the challenge test set as a benchmark, our pipelined approach achieves state-of-the art performance for all three sub-tasks, in both automated and manual evaluation.


\begin{figure*}[t]
    \centering
    \includegraphics[width=0.9\textwidth]{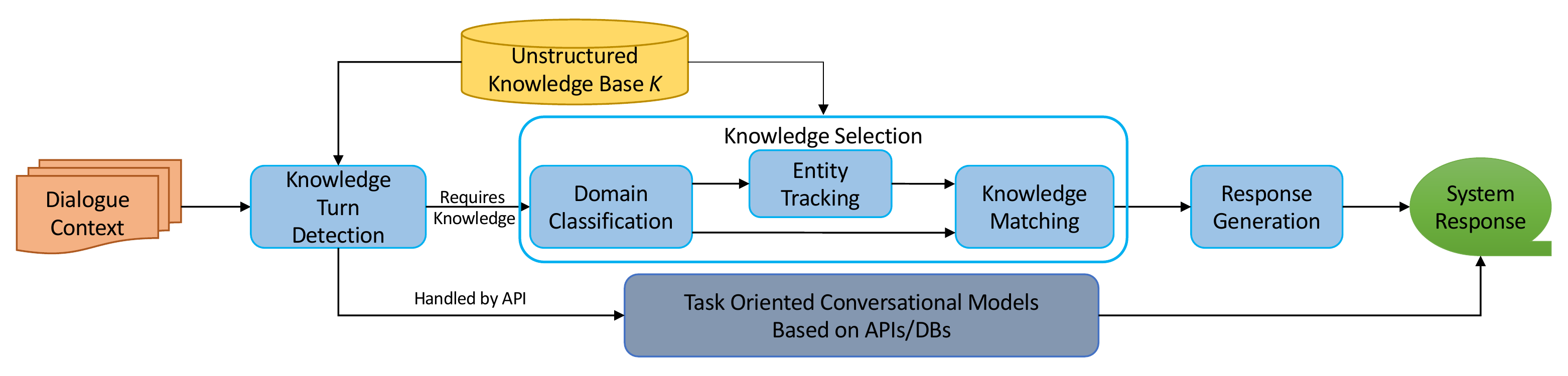}
    \caption{Task formulation and architecture of our knowledge-grounded dialog system.}
    \label{fig:method-schema}
\end{figure*}



\section{Approach}
Our approach to task-oriented conversation modeling with unstructured knowledge access~\citep{kim-etal-2020-beyond} includes three successive sub-tasks, as illustrated in Figure \ref{fig:method-schema}. First, knowledge-seeking turn detection aims to identify user requests that are beyond the coverage of the task API. Then, for detected queries, knowledge selection aims to find the most appropriate knowledge that can address the user queries from a provided knowledge base. Finally, knowledge-grounded response generation produces a response given the dialogue history and selected knowledge. 

DSTC9 Track 1~\citep{kim2020beyond} organizers provided a baseline system that adopted the fine-tuned GPT2-small~\citep{radford2019language} for all three sub-tasks. The winning teams (Team 19 and Team 3) extensively utilized ensembling strategies to boost the performance of their submissions~\citep{he2021learning,tang2021radge,mi2021towards}.
We follow the pipelined architecture of the baseline system, but made innovations and improvements for each sub-task, outlined in detail below.

\subsection{Knowledge-seeking Turn Detection}

We treat knowledge-seeking turn detection as a binary classification task, given the dialogue context as the input, and fine-tuned a pre-trained language model for this purpose. The knowledge provided in the knowledge base constitutes a set of FAQs. We augmented the available training sets by treating all questions in the knowledge base as new potential user queries. 
Furthermore, for all questions in this augmentation that contain an entity name, we created a new question by replacing this entity name with ``it''. In this way, we obtained 13,668 additional data samples. 
In contrast to the baseline, we found that replacing GPT2-small with RoBERTa-Large~\citep{DBLP:journals/corr/abs-1907-11692} improved the performance. The other changes we made include feeding only the last user utterance instead of the whole dialogue context into the model and fine-tuning the decision threshold $t_{ktd}$ (when the inferred probability score $p>t_{ktd}$, the prediction is positive, otherwise negative) to optimize the F1 score on the validation set, both of which helped achieve better performance.
    
\subsection{Knowledge Selection}

For knowledge selection, the baseline system predicts the relevance between a given dialogue context and every candidate in the whole knowledge base, which is very time-consuming especially when the size of knowledge base is substantially expanded. Instead, we propose a hierarchical filtering method to narrow down the candidate search space. Our proposed knowledge selection pipeline includes the following three modules: domain classification, entity tracking, and knowledge matching, as illustrated in Figure \ref{fig:method-schema}. Specifications of each module are detailed below.

\subsubsection{Domain Classification}
\label{sec:domain_classification}

In multi-domain conversations, if the system knows what domain a given turn belongs to, the search space for knowledge selection can be greatly reduced by taking the domain-specific knowledge only.
The DSTC9 Track 1 data includes the augmented turns for ``Train'', ``Taxi'', ``Hotel'', and ``Restaurant'' domains in its training set, where the first two domains have domain-level knowledge only, while the others can be further subdivided for each entity-specific knowledge. To improve the generalizability of our filtering mechanism for unseen domains, we merged the domains which require further entity-level analysis into an ``Others'' class and defined this task as a three-way classification:  \{``Train'', ``Taxi'', and ``Others''\}.

We implemented a domain classifier by fine-tuning the RoBERTa-Large model which takes the whole dialogue context and outputs a domain label.
Considering that a new domain (i.e., ``Attraction'') is introduced in the test set, we augmented the training data with 3,350 additional samples of the ``Attraction'' domain, which were obtained from the MultiWOZ 2.1~\cite{eric-etal-2020-multiwoz}, the source of the DSTC9 Track 1 data (all augmented samples are labeled as ``Others''). More specifically, we first find out those dialogues for ``Attraction'' in the training set of the MultiWOZ 2.1 dataset (this dataset contains seven domains including ``Attraction'') by selecting dialogue turns that contain ``Attraction'' related slots. We then replace the original ``Attraction'' related slots with entities of the ``Attraction'' domain in the knowledge base $K$. Meanwhile we replace the last user utterances in the dialogues with the knowledge questions that belong to the replaced new entities. Table \ref{table:data-augmentation-domain-classification-sample} gives one example for explanation. In this example, we replace the original entity of ``funky fun house'' with a new entity of ``California Academy of Science'' randomly selected from the ``Attraction'' domain of the knowledge base. Besides, we replace the original last user utterance with a knowledge question randomly selected from the FAQs of this new entity ``California Academy of Science''. 

\begin{table*}[t]
\small
\centering
\resizebox{0.9\textwidth}{!}{
\begin{tabular}{p{0.05\textwidth}p{0.47\textwidth}p{0.47\textwidth}}
\toprule
 Speaker & Original Dialogue & New Dialogue \\ \midrule
 User & I was hoping to see local places while in Cambridge. Some entertainment would be great. & I was hoping to see local places while in Cambridge. Some entertainment would be great.  \\
 Agent & I got 5 options. which side is okay for you? & I got 5 options. which side is okay for you? \\
 User & It doesn't matter. Can I have the address of a good one? & It doesn't matter. Can I have the address of a good one? \\
 Agent & How about \textcolor{red}{funky fun house}, they are located at 8 mercers row, mercers row industrial estate. & How about \textcolor{blue}{California Academy of Sciences}, they are located at 8 mercers row, mercers row industrial estate. \\
 User        & \textcolor{red}{Could I also get the phone number and postcode?} &    \textcolor{blue}{Is WiFi available?}                       \\ \bottomrule
\end{tabular}}
\caption{An example of data augmentation for domain classification. The left dialogue is the original dialogue from the MultiWOZ 2.1 dataset while the right one is synthesized by replacing the original entity and last user utterance highlighted by red with a new entity and knowledge question from the knowledge base highlighted by blue.}
\label{table:data-augmentation-domain-classification-sample}
\end{table*}
\subsubsection{Entity Tracking}
\label{sec:entity_exraction}

Once the domain classifier predicts the 'Others' label for a given turn, the entity tracking module is executed to detect the entities mentioned in the dialogue context and align them to the entity-level candidates in the knowledge base.
We adopt an unsupervised approach based on fuzzy n-gram matching whose details can be referred to Section A.2 of the Supplementary Material. After extracting these entities, we determined the character-level start position of each entity in the dialogue context and selected the last three mentioned entities as the output of this module. 

\subsubsection{Knowledge Matching}

The knowledge matching module receives a list of knowledge candidates and ranks them in terms of relevance to the input dialogue context.
We concatenated the dialogue context, domain/entity name, and each knowledge snippet into a long sequence, which is then sent to the fine-tuned RoBERTa-Large model to get a relevance score.

To train the model, we adopted Hinge loss, which was reported to perform better for the ranking problems~\citep{wang2014learning,NEURIPS2018_42998cf3} than Cross-entropy loss used in the baseline system.
For each positive instance, we drew four negative samples, each of which is randomly selected from one of four sources: 1) the whole knowledge base, 2) the knowledge snippets in the ground truth domain, 3) the knowledge snippets of the ground truth entity, and 4) the knowledge snippets of other entities mentioned in the same dialogue.
In the execution time, we fed the knowledge candidates filtered by the predicted domain and entity from Section~\ref{sec:domain_classification} and ~\ref{sec:entity_exraction}, repectively.
Then, the module outputs a list of the candidates ranked by relevance score.



\begin{table}[!hptb]
\small
\centering
\begin{tabular}{lccc}
\toprule
                      & Precision       & Recall       & F1      \\ \midrule
Our proposed model & 0.9920 & 0.9344 & 0.9623 \\
+ data augmentation & 0.9903 & \underline{0.9833} & \underline{0.9868} \\
\midrule
\multicolumn{4}{l}{\textbf{\textit{DSTC9 Track 1 Systems:}}} \\
\hspace{0.2cm}Baseline &   0.9933   &    0.9021    & 0.9455       \\
\hspace{0.2cm}Team $17$$^\dagger$ & \underline{0.9933} & 0.9748 & 0.9839 \\
\hspace{0.2cm}Team $3$$^\ddagger$ & \textbf{0.9964} & \textbf{0.9859} & \textbf{0.9911}   \\ \bottomrule
\end{tabular}
\caption{Test results on task 1: knowledge-seeking turn detection. $^\dagger$ and $^\ddagger$ denote the best DSTC9 Track 1 systems with a single model and model ensemble, respectively. Overall highest scores are made bold while highest scores for single models are underlined.}
\label{table:sub-task-1}
\end{table}


\begin{table}[!hptb]
\small
\centering
\begin{tabular}{lccc}
\toprule
                      & MRR@5       & Recall@1       & Recall@5       \\ \midrule
Our proposed model &  \underline{0.9461} & \textbf{\underline{0.9251}} & \underline{0.9702}        \\ \midrule
\multicolumn{4}{l}{\textbf{\textit{DSTC9 Track 1 Systems:}}} \\
\hspace{0.2cm}Baseline & 0.7263 & 0.6201 & 0.8772       \\
\hspace{0.2cm}Team $7$$^\dagger$ & 0.9309 & 0.8988 & 0.9666\\
\hspace{0.2cm}Team $19$$^\ddagger$&  \textbf{0.9504}  &	0.9235 &	\textbf{0.9840}   \\ \bottomrule
\end{tabular}
\caption{Test results on task 2: knowledge selection.}
\label{table:sub-task-2}
\end{table}

\begin{table*}[t]
\small
\centering
\begin{tabular}{lcccccccc}
\toprule
              & BLEU-1 & BLEU-2 & BLEU-3 & BLEU-4 & METEOR & ROUGE-1 & ROUGE-2 & ROUGE-L \\ \midrule
              \textbf{\textit{Our Systems:}} \\
\hspace{0.2cm}BART-Large &  0.3743      & 0.2428       & 0.1620       &    0.1098    & 0.3869       & 0.4163        & 0.1992        & 0.3639        \\
\hspace{0.2cm}T5-Base &  0.3575      & 0.2432       & 0.1685       &    0.1155    & \underline{\textbf{0.4379}}       & 0.4139        & 0.2103 & 0.3536        \\
\hspace{0.2cm}Pegasus-Large &   \underline{0.3808} & 0.2531 &  0.1727 &    \underline{\textbf{0.1192}}    & 0.4013       & \underline{0.4237}        & 0.2099         & 0.3656        \\ \midrule
\multicolumn{9}{l}{\textbf{\textit{DSTC9 Track 1 Systems:}}} \\
\hspace{0.2cm}Baseline &  0.3031 &	0.1732 &	0.1005 &	0.0655 &	0.2983 &	0.3386 &	0.1364 &	0.3039        \\
\hspace{0.2cm}Team $15$$^\dagger$ & 0.3779 & \underline{0.2532} & \underline{\textbf{0.1731}} & 0.1175 & 0.3931 & 0.4204 & \underline{0.2113} & \underline{0.3765} \\
\hspace{0.2cm}Team $3$$^\ddagger$&   \textbf{0.3864} &	\textbf{0.2539} &	0.1692 &	0.1190 &	0.3914 &	\textbf{0.4332} &	\textbf{0.2115} &	\textbf{0.3885}  \\ \bottomrule
\end{tabular}
\caption{Test results on task 3: knowledge grounded response generation.}
\label{table:sub-task-3}
\vspace{-2mm}
\end{table*}

\subsection{Response Generation}

For response generation, we compared the following three pre-trained sequence-to-sequence (seq2seq) models: T5-Base~\citep{JMLR:v21:20-074}, BART-Large~\citep{lewis-etal-2020-bart}, and Pegasus-Large~\citep{pmlr-v119-zhang20ae}.
Each model inputs a concatenated sequence of the whole dialogue context and the knowledge answer and then outputs a response. The ground-truth knowledge answer is used in the training phase, while the top-1 candidate from the knowledge selection result is used in the test phase.

\section{Experiments and Results}


We used the same data split and evaluation metrics as the official DSTC9 Track 1 challenge.
All model training and dataset details are summarized in the Section B of the Supplementary Material.
\subsection{Knowledge Seeking Turn Detection}
Table \ref{table:sub-task-1} compares the knowledge seeking turn detection performance between our proposed models and the best single model and ensemble-based systems from the DSTC9 Track 1 official results.\footnote{There are up to five entries submitted by each team in the competition and we report only the best entries by a single model and ensemble-based systems.}
The results show that our proposed data augmentation method helped to improve the recall of our detection model and led to the highest F1 score among all the single models in the challenge.




\subsection{Knowledge Selection}
Our domain classification and entity tracking modules achieved 99.5\% in accuracy and 97.5\% in recall, respectively.
The data augmentation method helped to improve the domain classification accuracy from 97.1\% to 99.5\%.

Table \ref{table:sub-task-2} summarizes the knowledge selection performance of our system based on the proposed hierarchical filtering mechanism using the results from both domain classification and entity tracking modules.
Our proposed system outperformed the challenge baseline in all three metrics with a largely reduced execution time from more than 20 hours by the baseline to less than half an hour to process the whole test set with a single V100 GPU.

Compared with the best knowledge selection results from the challenge, our model achieved higher performances than the best single model-based system in all metrics, and even surpassed the best ensemble model in recall@1.
To be noted, recall@1 is the most important metric, since the response generation is grounded on only the top-1 result from knowledge selection.

\subsubsection{Ablation Study}

First of all, Table \ref{table:ablation-hinge-loss} summarizes the ablation results by imposing two kinds of changes based on our full knowledge matching model: instead of concatenating the dialogue context, domain name, entity name, and knowledge question and answer pair as the input to the model, we only concatenate the dialogue context and knowledge question and answer pair (w/o entity names); we replace the Hinge loss with Cross-entropy loss (w/o Hinge Loss). To be noted, we should pay more attention to the Recall@1 score in the Table \ref{table:ablation-hinge-loss}, which is the most important metric. And we can see that adding the domain and entity names are beneficial and the use of Hinge loss for optimization is better than Cross-entropy for this ranking problem.

As above-mentioned, for training the knowledge matching module, we need to sample several negative samples for each position sample and instead of using only one negative sampling strategy, we used a mixed strategy. More specifically, for sampling each negative sample, we randomly adopted one of the following four strategies:

\begin{enumerate}
    \itemsep-0.3em  
    \item Randomly select from all knowledge snippets;
    \item Randomly select from the knowledge snippets of entities that are the in the same domain as the ground truth one (i.e., the entity of the positive sample);
    \item Randomly select from the knowledge snippets of the ground truth entity;
    \item Randomly select from the knowledge snippets of entities that are mentioned in the same dialogue as the ground truth one.
\end{enumerate}

Each strategy $i\in \{1,2,3,4\}$ is sampled at a certain sampling ratio $p_{ns}^i$. We tuned this sampling ratio by trying several combinations, and the results are summarized in Table \ref{table:ablation-sampling-ratio}. From it, we can see that: (1) Strategy 4 is the most effective among all four ones; (2) Mixing four strategies is better than using only one of them; (3) Allocating higher ratio to strategy 4 is better than uniform ratios for every strategy.

\begin{table}[!hptb]
\centering
\small
\begin{tabular}{lrrr}
\toprule
Settings & MRR@5 & Recall@1 & Recall@5 \\ \midrule
Original model  & \textbf{0.9811}       & \textbf{0.9693}            & \textbf{0.9936}          \\
\hspace{0.2cm}w/o entity names       & 0.9788       & 0.9656         & 0.9933          \\
\hspace{0.2cm}w/o Hinge Loss       &  0.9734     & 0.9613          & 0.9905         \\ \bottomrule
\end{tabular}
\caption{Ablation study of the knowledge matching module for knowledge selection by removing entities and hinge loss. Scores are reported on the validation set.}
\label{table:ablation-hinge-loss}
\end{table}

\begin{table}[!hptb]
\centering
\small
\resizebox{0.99\columnwidth}{!}{
\begin{tabular}{lrrr}
\toprule
\textbf{Sampling ratios} & \textbf{MRR@5} & \textbf{Recall@1} & \textbf{Recall@5} \\ \midrule
\textbf{\textit{Original model}} \\
\hspace{0.2cm}[0.1,0.1,0.1,0.7]  & \textbf{0.9811}       & \textbf{0.9693}            & \textbf{0.9936}          \\ \midrule
\hspace{0.2cm}[0.25,0.25,0.25,0.25]       &    0.9761 & 0.9615 & 0.9929       \\
\hspace{0.2cm}[1.0,0.0,0.0,0.0]       &     0.9712 & 0.9514 & 0.9933      \\
\hspace{0.2cm}[0.0,1.0,0.0,0.0]       &  0.9559 & 0.9248 & 0.9906           \\ 
\hspace{0.2cm}[0.0,0.0,1.0,0.0]       &  0.9728 & 0.9540 & 0.9933           \\
\hspace{0.2cm}[0.0,0.0,0.0,1.0]       & 0.9751 & 0.9596 & 0.9929            \\ 
\bottomrule
\end{tabular}}
\caption{Ablation study of the knowledge matching module for knowledge selection by tuning the mixed negative sampling ratio. Scores are reported on the validation set. The sampling ratio is represented in the format of $[p_{ns}^1,p_{ns}^2,p_{ns}^3,p_{ns}^4]$.}
\label{table:ablation-sampling-ratio}
\end{table}



\subsection{Response Generation}
Table \ref{table:sub-task-3} summarizes the automated evaluation results for the generated responses with different seq2seq models.
Our fine-tuned T5-Base model achieved lower BLEU scores than BART-Large and Pegasus-Large, while its METEOR score is substantially higher than the others.
Note that our generation system does not perform any model ensemble, and it surpasses the best single system in the DSTC9 Track 1 for half of the metrics. 

Following the official evaluation protocol in the challenge, we performed human evaluation to compare our system with the top systems from the challenge\footnote{https://github.com/alexa/alexa-with-dstc9-track1-dataset/tree/master/results}, as shown in Table \ref{table:sub-task-3-human-eval}.
Specifically, we hired three crowd-workers for each instance, asked them to score each system output in terms of its ``accuracy'' and ``appropriateness'' in five point Likert scale, and reported the averaged scores. 
We have three findings:
(1) T5 achieves higher accuracy, while Pegasus is slightly better for appropriateness;
(2) our systems generates more accurate responses than the top DSTC9 systems, while the appropriateness scores is comparable (confirmed by significance testing in Section C.2 of Supplementary Material);
(3) the final average scores of our systems rank the highest.
We present several examples of the generated responses by our system compared against the baseline and top 2 systems in Section C.3 of Supplementary Material.

\begin{table}[!htpb]
\small
\centering
\begin{tabular}{lccc}
\toprule
Systems       & Accuracy & Appropriateness & Average \\ \midrule
\textbf{\textit{Our Systems:}} \\
T5-Base       &  \textbf{4.5994$^\ast$}	& 4.4572$^\dagger$ &	\textbf{4.5283}$^\ast$          \\
Pegasus-Large &          4.5451$^\dagger$ &	4.4591$^\dagger$ &	4.5021$^\dagger$             \\ \midrule
\multicolumn{3}{l}{\textbf{\textit{DSTC9 Track 1 Systems (Top-2):}}}  \\
Team 19        &  4.4979  &	\textbf{4.4698} &	4.4838 \\
 & (4.3917) & (4.3922) & (4.3920) \\
Team 3       & 4.4524 & 4.4064 &	4.4294 \\ 
 & (4.3480) & (4.3634) & (4.3557) \\
\bottomrule
\end{tabular}
\caption{Human evaluation results of the test set for response generation.
Numbers within the parentheses are official scores from DSCT9~\citep{kim2020beyond}.
The symbol $^\ast$ means our score is significantly higher than the best previous system while $^\dagger$ means our score is not significantly different from the best previous system, according to paired t-test with $p<0.05$.}
\label{table:sub-task-3-human-eval}
\end{table}

\section{Conclusions}

In this work, we propose a comprehensive system to enable the task-orientated dialogue models to answer user queries that are out of the scope of APIs. We significantly improved the system's capability of finding the most relevant knowledge snippets, consequently providing excellent responses by introducing a novel data augmentation method, incorporating domain and entity identification modules for knowledge selection, and utilizing mixed negative sampling. To demonstrate the efficacy of our approach, we benchmark our system on the DSTC9 Track 1 challenge dataset and report the state-of-the-art performance.

\bibliographystyle{acl_natbib}
\bibliography{acl2021}

\newpage

\appendix

\setcounter{table}{0}
\renewcommand\thetable{\Alph{section}.\arabic{table}}

\section{Methods}



\subsection{Entity Extraction}

Specifically, we first normalize the entity names in the knowledge base using a set of heuristic rules, such as replacing the punctuation ``\&'' with ``and''. Table \ref{table:entity-normalization} summarizes the full list of normalization rules and we give an example for each rule as illustration. Then we perform the fuzzy n-gram matching between an entity and a certain piece of dialogue context. For example of an entity of ``Alexander Bed and Breakfast'', it is a four-gram, therefore we extract all four-grams from the dialogue context and match each of them against it. And the process of matching is to first find out the longest contiguous matching sub-sequence and then calculate the matching ratio by the equation of $2M/T$, where $M$ is the length of the matched sub-sequence while $T$ is the total length of the two n-grams to be matched.\footnote{https://towardsdatascience.com/sequencematcher-in-python-6b1e6f3915fc} If this ratio is higher than 0.95, we deem this pair of n-grams as matched. In this way, we can find out which entities in the knowledge base are mentioned in a certain dialogue.

\begin{table*}[t]
\small
\centering
\resizebox{0.9\textwidth}{!}{
\begin{tabular}{p{0.6\textwidth}p{0.42\textwidth}}
\toprule
Normalization rules & Examples \\ \midrule
Replace the punctuation ``\&'' with ``and'' &  Bay Subs \& Deli $\rightarrow$ Bay Subs and Deli \\
If the entity contains any symbol of `` - '', ``, '' or ``/'', split this entity by this symbol and remove the second part & Hard Knox Cafe - Potrero Hill $\rightarrow$ Hard Knox Cafe \\
Replace ``guesthouse'' with ``guest house'' & ARBURY LODGE GUESTHOUSE $\rightarrow$ ARBURY LODGE GUEST HOUSE \\
If the entity contains a place name such as ``Fisherman's Wharf'' and ``San Francisco'' in the end, remove it (since the entities in the knowledge base do not contain these place names) & Bay Bridge Inn San Francisco $\rightarrow$ Bay Bridge Inn \\
Replace numbers with corresponding English words & Pho Huynh Hiep 2 $\rightarrow$ Pho Huynh Hiep Two                      \\ \bottomrule
\end{tabular}}
\caption{Entity normalization rules and the corresponding examples.}
\label{table:entity-normalization}
\end{table*}

\section{Experiments}

\begin{table*}[t]
\small
\centering
\resizebox{0.99\textwidth}{!}{
\begin{tabular}{p{0.02\textwidth}p{0.05\textwidth}p{0.51\textwidth}p{0.41\textwidth}}
\toprule
Turn & Speaker & Utterance & Sampled Knowledge Snippets from FAQs \\ \midrule
1     & User        & I'm looking for a place to stay in the south of town. It doesn't need to have free parking.          &                            \\
2     & Agent        &  There are 4 hotels that are in the area you are looking for. Would you prefer a 3 or 4 star rated hotel?         &                            \\
3     & User        &   I don't care about the star rating as long as it's expensive.        &                            \\
4     & Agent        &  The \textbf{\textit{Lensfield Hotel}} is the only expensive hotel in the south area. Would you like any more information on this location?         &                            \\
5     & User        &  I'm interested in knowing, do they have a workout facility on the premises?         &                            \\
6     & Agent        &  There are both a fitness center and gym available on the premises. Does this sound ok?         & Q1: Do you have room service for your guests?\\
& & & A1: Yes, the Lensfield Hotel provides room services.\\
& & & \textbf{Q2: Is there a gym available at your location?} \\
& & & \textbf{A2: There is both a fitness center and gym available on the premises.} \\
& & & Q3: Can I bring my dog? \\
& & & A3: Pets are not allowed at the Lensfield Hotel. \\
7     & User        &  That is perfect can you book that for me please.         &                            \\
8     & Agent        &   The Lensfield Hotel is located in the South. It has a 3 star rating and is expensive. There is free parking and internet. I have booked it for you.        &                            \\
9     & User        & Great, thank you!          &                           \\ \bottomrule
\end{tabular}}
\caption{Examples of task-oriented conversations with unstructured knowledge access. Three sampled FAQ pairs for the entity ``Lensfield Hotel'' are listed in the rightmost column for turn 5 which is beyond the coverage of API and needs external knowledge support. The most appropriate FAQ pair to address turn 5 is highlighted in bold font.}
\label{table:data-sample}
\end{table*}

\subsection{Data Samples \& Statistics}

Table \ref{table:data-sample} shows an example conversation with unstructured knowledge access. The user utterance at turn $t=5$ requests the information about the gym facility, which is out of the coverage of the structured domain APIs. However, the relevant knowledge contents can be found from the external sources as in the rightmost column which includes the sampled QA snippets from the FAQ lists for each corresponding entity within domains such as train, hotel, or restaurant. With access to these unstructured external knowledge sources, the agent manages to continue the conversation with no friction by selecting the most appropriate knowledge.

The data statistics are summarized in Table \ref{table:data-stats}.\footnote{Data can be downloaded from: https://github.com/alexa/alexa-with-dstc9-track1-dataset}  The  main  data  is  an  augmented  version  of  MultiWOZ  2.1 that includes newly introduced knowledge-seeking turns in the MultiWOZ conversations. A total of 22,834 utterance pairs were newly collected based on 2,900 knowledge candidates from the FAQ webpages about the  domains  and  the  entities  in  MultiWOZ  databases. To be noted, for the test set, other conversations collected from scratch about touristic information for San Francisco are added. To  evaluate  the  generalizability  of  models,  the  new conversations cover knowledge, locale and domains that are unseen from the train and validation data sets. In addition, this test set includes not only written conversations, but also spoken dialogues to evaluate system performance across different modalities.

\begin{table}[!htpb]
\small
\centering
\resizebox{0.99\columnwidth}{!}{
\begin{tabular}{llrrr}
\toprule
Split & Source & \# dialogues & \# samples & \# knowledge seeking turns  \\ \midrule
Train & MultiWOZ      & 7,190 & 71,348 & 19,184              \\ \midrule
Valid & MultiWOZ     &    1,000            & 9,663 & 2,673               \\ \midrule
\multirow{3}{*}{Test} & MultiWOZ    &   977 & 2,084 & 977\\
 & SF Written      & 900 &1,834 & 900  \\ 
& SF Spoken      & 107 & 263 & 104 \\ \bottomrule
\end{tabular}}
\caption{Statistics of the data divided into training, validation, and test sets. The test set contains three sources of samples: MultiWOZ, San Francisco tourism in written English, and San Francisco tourism in spoken English, which is different from train and validation sets.}
\label{table:data-stats}
\end{table}

Table \ref{table:knowledge-stats} gives the statistics of the knowledge base, which is a collection of frequently asked questions (FAQs). To be noted, there are no entities for the ``Train'' and ``Taxi'' domains while for ``Hotel'', ``Restaurant'', and ``Attraction'' domains, each entity has its corresponding list of FAQ pairs. Besides, the knowledge base for the test set covers the train \& validation sets and is further expanded by adding one more domain of ``Attraction'' and more entities.

\begin{table}[!htpb]
\small
\centering
\resizebox{0.99\columnwidth}{!}{
\begin{tabular}{lrrrr}
\toprule
\multirow{2}{*}{Domain} & \multicolumn{2}{c}{Train \& Val} & \multicolumn{2}{c}{Test}  \\
                        & \# Entities    & \# Snippets   & \# Entities & \# Snippets \\ \midrule
Train                   &  --              &  26             &  --           &  26           \\
Taxi                    &  --              &  5             &   --          &  5            \\
Hotel                   &  33              & 1,219              &  178           &  4,346           \\
Restaurant              &  110              & 1,650              &  391           & 7,155            \\
Attraction              &  --              &  --             &  97           &  507          \\ \midrule 
Total                   &  143             & 2,900              & 666              &    12,309       \\\bottomrule
\end{tabular}}
\caption{Statistics of the knowledge base (the list of FAQs). ``Train'' and ``Taxi'' domains do not have any entities and there is no ``Attraction'' domain for the knowledge base in train and validation sets.}
\label{table:knowledge-stats}
\end{table}

\subsection{Experimental Details}

We implemented our proposed system based on the DSTC9 Track 1 baseline provided by \citet{kim2020beyond} and the transformers library~\citep{wolf-etal-2020-transformers}. For all sub-tasks, the maximum sequence length for the dialogue context and the knowledge snippet is both 128. For the knowledge seeking turn detection sub-task, the model is fine-tuned for 5 epochs with the batch size of 16, while for other sub-tasks, 8 epochs and the batch size of 4 are used. A model checkpoint is saved after each epoch, and the best checkpoint is picked based on the validation results. For decoding process of the response generation model, we replaced the nucleus sampling in the baseline to beam search (beam width is 5), which achieved higher performances in the validation set.

\section{Results}

\begin{table*}[t]
\small
\centering
\resizebox{0.9\textwidth}{!}{
\begin{tabular}{lcccccc}
\toprule
 & \multicolumn{2}{c}{Accuracy} & \multicolumn{2}{c}{Appropriateness} & \multicolumn{2}{c}{Average} \\
 & T5-Base           &     Pegasus-Large      & T5-Base           &   Pegasus-Large       & T5-Base           &    Pegasus-Large      \\ \midrule
Team 19 &           $1.86\times 10^{-6}$        & $1.45\times 10^{-1}$          &           $6.07\times 10^{-1}$        & $5.56\times 10^{-1}$          &           $1.96\times 10^{-3}$        & $4.25\times 10^{-1}$          \\
Team 3 &           $1.21\times 10^{-16}$         & $8.01\times 10^{-7}$          &           $2.89\times 10^{-4}$         & $5.68\times 10^{-5}$          &           $8.31\times 10^{-15}$         & $2.48\times 10^{-8}$         \\ \bottomrule
\end{tabular}}
\caption{Significance Testing p-values between our systems and top-2 systems in the DSTC9 competition for the accuracy, appropriateness, and average score of response generation. Paired t-test is used.}
\label{table:significance-testing}
\end{table*}

\begin{table}[!htbp]
\small
\centering
\resizebox{0.99\columnwidth}{!}{
\begin{tabular}{p{0.09\textwidth}p{0.42\textwidth}}
\toprule
                 & \textbf{Example}  \\ \midrule
\textbf{Dialogues}         &  \textbf{User:} I need a train to Broxbourne, that arrives by 20:30.\newline \textbf{Agent:} Certainly. Can I assume you're leaving from Cambridge?\newline \textbf{User:} Yes, that is correct.\newline \textbf{Agent:} There is a train leaving at 5:01, would you like a seat?\newline \textbf{User:} Is that leaving on Sunday?\newline \textbf{Agent:} I have the TR8991 leaving at 19:01 and arriving by 20:01. Would you like me to book that for you?\newline \textbf{User:} Not yet but I also need a restaurant that serve mediterranean food and should be in the centre of town.\newline \textbf{Agent:} I have a few restaurants that fit your criteria. The Gardenia, La Mimosa, and Shiraz Restaurant. Which one are you interested in hearing about?\newline \textbf{User:} Tell me about The Gardenia.\newline \textbf{Agent:} It's a cheap restaurant located at 2 Rose Crescent City Centre. Phone number and postcode are 01223356354 and cb23ll. Do you need anything else?\newline \textbf{User:} Does the restaurant offer live music on the weekend?\newline \textbf{Agent:} Sorry, there is no live music at The Gardenia. What else can I help you with?\newline \textbf{User:} Does the train have a data limit for wifi usage?                  \\ \midrule
\textbf{Knowledge Answer} &  Data is limited to 50MB per day with no option of additional data.                \\ \midrule
\multicolumn{2}{l}{\textbf{\textit{Our systems:}}}  \\
\hspace{0.2cm}\textbf{T5-Base}          & Data is limited to 50MB per day with no option of additional data. Is there anything else I can do for you today or would you like to make a reservation?                  \\
\hspace{0.2cm}\textbf{Pegasus-Large}    & Data is limited to 50MB per day with no option of additional data. Is there anything else I can help you with?                  \\ \midrule
\multicolumn{2}{l}{\textbf{\textit{Top-2 submitted systems:}}}  \\
\hspace{0.2cm}\textbf{Team 3}           & No, there is no additional data available to purchase. Anything else I can do for you?                   \\
\hspace{0.2cm}\textbf{Team 19}          & No, the train does not have a data limit for wifi usage. Anything else I can do for you?                 \\ \bottomrule
\end{tabular}
}
\caption{Qualitative comparison between our system with previous strong competitors. Knowledge answer is the answer part of the ground truth knowledge snippet. We are comparing against the top-2 systems submitted to the DSTC9 competition.}
\label{table:qualitative-comparison}
\end{table}

\subsection{Significance Testing for Human Evaluation}

Since those scores of human evaluation for response generation are quite close to each other, we resort to significance testing to confirm our system's superior performance. Table \ref{table:significance-testing} summarizes the significance testing p-value between our systems and the top-2 submitted systems in the DSTC9 challenge for the accuracy, appropriateness, and average scores, respectively. From it, we can see that T5-Base is significantly higher than the competing systems in terms of accuracy ($p<0.05$). Besides, T5-Base and Pegasus-Large are comparable to the best previous system in terms of appropriateness. Finally, with regards to the average score, our T5-Base significantly rivals the previous best system. 

\subsection{Qualitative Examples of Responses}

Table \ref{table:qualitative-comparison} gives one qualitative example to compare our system's responses against those of the top-2 submitted systems in the DSTC9 competition (i.e., Team 3 and 19)\footnote{https://github.com/alexa/alexa-with-dstc9-track1-dataset/tree/master/results}. Overall, we can see that our system's responses are more accurate. Taking the example in Table \ref{table:qualitative-comparison}, our responses can exactly answer the user query and it is strictly aligning with the ground truth knowledge, while the response from Team 19 is totally wrong and that from Team 3 does not address the user query at all.

\end{document}